\documentclass[10pt,twocolumn,letterpaper]{article}

\usepackage[pagenumbers]{humanrig} 
\usepackage{multirow}
\usepackage{amsmath}
\usepackage{algorithm}
\usepackage{algpseudocode}
\usepackage{amsfonts}
\usepackage{pifont}

\definecolor{cvprblue}{rgb}{0.21,0.49,0.74}
\usepackage[pagebackref,breaklinks,colorlinks,allcolors=cvprblue]{hyperref}

\title{HumanRig: Learning Automatic Rigging for Humanoid Character in a Large Scale Dataset}

\author{Zedong Chu, Feng Xiong, Meiduo Liu, Jinzhi Zhang, Mingqi Shao, Zhaoxu Sun, Di Wang, Mu Xu\\
AMAP, Alibaba \\
{ \{chuzedong.czd, xf250971, liumeiduo.l, wushou.zjz,} \\
{ mingqi.smq, szx430629, wd426042, xumu.xm\}@alibaba-inc.com~}}

\begin{document}
\maketitle

\begin{abstract}
With the rapid evolution of 3D generation algorithms, the cost of producing 3D humanoid character models has plummeted, yet the field is impeded by the lack of a comprehensive dataset for automatic rigging—a pivotal step in character animation. Addressing this gap, we present HumanRig, the first large-scale dataset specifically designed for 3D humanoid character rigging, encompassing 11,434 meticulously curated T-posed meshes adhered to a uniform skeleton topology. Capitalizing on this dataset, we introduce an innovative, data-driven automatic rigging framework, which overcomes the limitations of GNN-based methods in handling complex AI-generated meshes. Our approach integrates a Prior-Guided Skeleton Estimator (PGSE) module, which uses 2D skeleton joints to provide a preliminary 3D skeleton, and a Mesh-Skeleton Mutual Attention Network (MSMAN) that fuses skeleton features with 3D mesh features extracted by a U-shaped point transformer. This enables a coarse-to-fine 3D skeleton joint regression and a robust skinning estimation, surpassing previous methods in quality and versatility. This work not only remedies the dataset deficiency in rigging research but also propels the animation industry towards more efficient and automated character rigging pipelines.
\end{abstract}

\section{Introduction}
\label{sec:intro}

\begin{figure}
  \centering
    \includegraphics[width=1.0\linewidth]{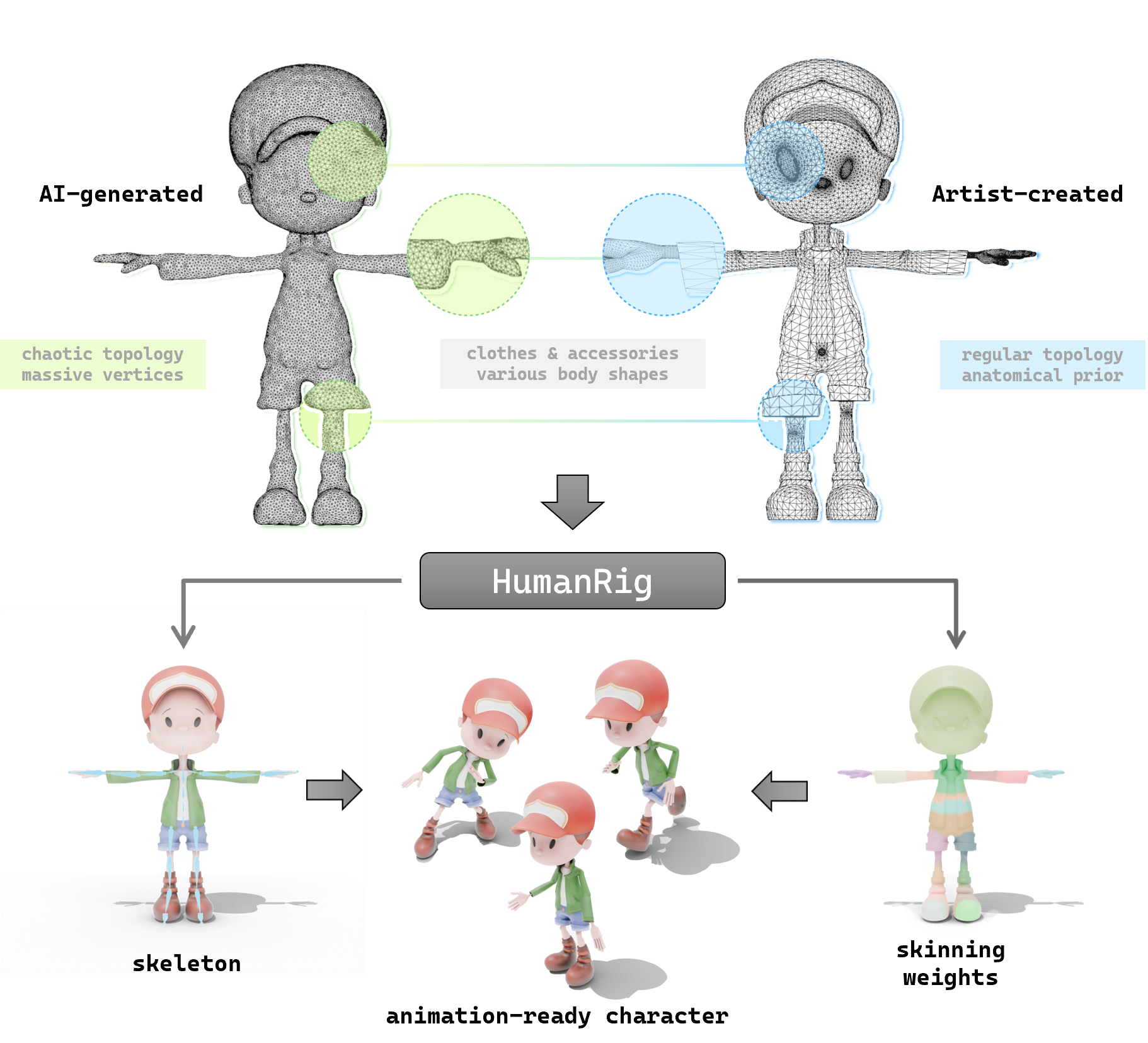}
    \caption{The AI-generated mesh and the artist-created one show distinct face topology distributions. Manual meshes often have varying vertex densities across body parts and special topologies near joints for better deformation, while AI meshes tend to have chaotic topologies which lack semantic information. Previous rigging methods can only handle simple artist-created meshes, while HumanRig deals well with both AI and manual ones, especially for those with complex clothes or accessories and irregular body shapes.}
    \label{fig:aicompare}
\end{figure}


The animation industry is undergoing a revolutionary change with the integration of machine learning and data-driven techniques, particularly in the realms of 3D modeling and character animation. A critical phase in bringing 3D humanoid characters to life is rigging, which entails skeleton construction and skinning to facilitate realistic motion. Traditionally, rigging is a labor-intensive task generally requiring skilled artists to manually construct skeleton joints and paint skinning weights. As content creation pipelines accelerate and grow in complexity, the need for automation in rigging has become more pressing than ever.

Bolstered by extensive 3D datasets, the field of AI-driven 3D model generation~\cite{DBLP:journals/corr/abs-2409-06322, xu2024instantmesh, wu2024unique3d, zhang2024clay, tochilkin2024triposr} has experienced significant advancements.
Yet, the advancement in automatic rigging remains constrained due to the lack of a comprehensive and standardized dataset, which is essential for training powerful and robust rigging models. 
Previous work, like \cite{xu2020rignet}, laid the cornerstone for earlier rigging methods but was limited by a small scale of 1,729 T-pose humanoid meshes with inconsistent skeleton topologies and a lack of meaningful joint labels, hindering their use in animation.
Studies such as \cite{li2021learning}, utilizing the SMPL dataset \cite{loper2015smpl} and its clothed version from \cite{bhatnagar2019mgn}, were restricted to realistic human body shapes, missing the character diversity needed for varied applications. This has prompted the development of our HumanRig.

We introduce HumanRig, standing out as the first large-scale dataset tailored for the task of rigging 3D humanoid character models, and offering 11,434 meticulously curated AI-generated humanoid meshes. All models are generated in T-pose and aligned with an industry-standard skeleton topology, which enables direct plug-and-play in standard animation engines. Unparalleled in scale and diversity, HumanRig features an extensive range of body proportions and character types, spanning from real persons to cartoon characters and humanoid animals. 

Utilizing the HumanRig dataset, we introduce an innovative automatic rigging framework that diverges from previous approaches~\cite{liu2019neuroskinning,xu2020rignet,pan2021heterskinnet,li2021learning,mosella2022skinningnet,ma2023tarig}, which primarily rely on Graph Neural Networks (GNNs)~\cite{wang2019dynamic,hanocka2019meshcnn} to learn rigging directly from 3D mesh geometric features. Such methods struggle with complex 3D models, particularly those generated by AI. The core modules of our method are: a) The Prior-Guided Skeleton Estimator (PGSE), which initializes a coarse skeleton using 2D priors projected into 3D space, significantly reducing the complexity of the rigging task. b) A U-shaped Point Transformer~\cite{zhao2021point} serving as our mesh encoder, disregarding edges to enhance rigging robustness on intricate meshes. c) The Mesh-Skeleton Mutual Attention Network (MSMAN) enriching mesh and skeleton features with mutual information within a high-level semantic space, enabling joint optimization of skeleton construction and skinning. 

Our results showcase the method's superior performance, delivering high-quality automatic rigging solutions that eclipse current state-of-the-art techniques. This contribution not only addresses the scarcity of large-scale datasets for humanoid rigging but also presents an automatic rigging solution set to revolutionize the animation industry. By simplifying the rigging process,  we pave the way for more efficient and automated character animation, unlocking new frontiers in creative expression and content creation.

\begin{figure*}[ht]
  \centering
   \includegraphics[width=0.9\linewidth]{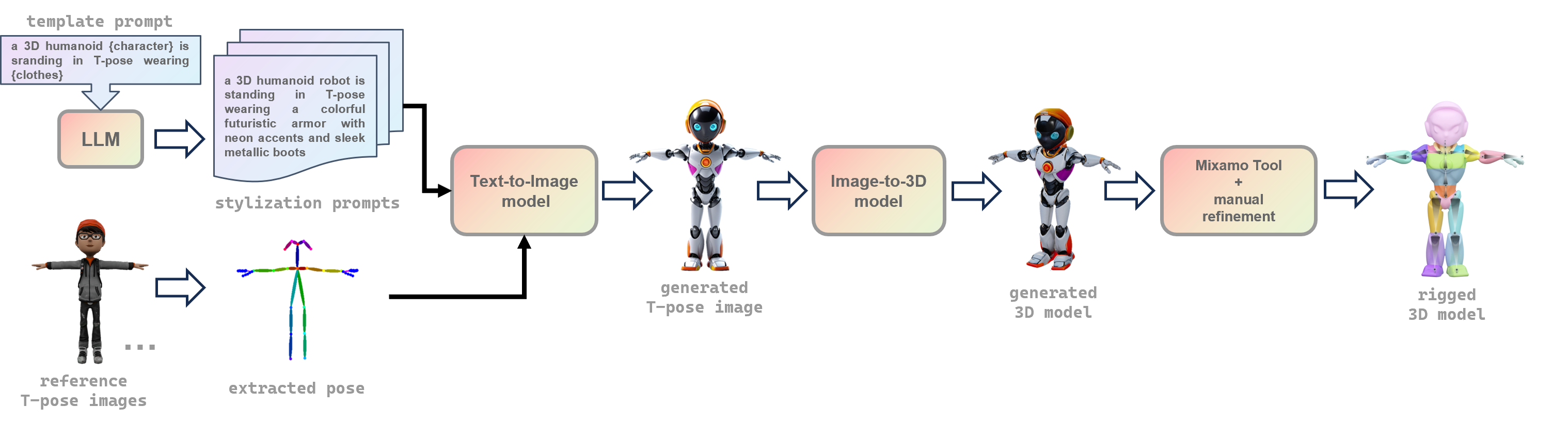}

   \caption{\textbf{Data Acquisition Pipeline} for our HumanRig dataset.}
   \label{fig:data pipeline}
\end{figure*}
\section{Related Work}

\subsection{3D Rigging Datasets}

The RigNetv1 dataset~\cite{xu2020rignet} has been instrumental in recent rigging research. It includes 2,703 rigged models, of which 1,729 are T-pose humanoid. Each character has between 1,000 and 5,000 vertices and an average of 25 skeleton joints. However, these models lack a standard skeleton topology, limiting the applicability of motion retargeting. SMPL~\cite{li2021learning} has gathered 3D data for real human body types using the same skeleton, but it is restricted to hairless bodies without clothes, and all samples share similar topologies and head-to-body ratios. Additionally, meshes in current datasets are manually crafted, which markedly differ from AI-generated meshes (See Fig.~\ref{fig:aicompare}). This discrepancy limits the generic learning of automatic rigging models for AI-generated humanoid characters.
Our HumanRig dataset, consisting of 11,434 AI-generated T-pose meshes with varied head-to-body ratios aligned to a standard Mixamo skeleton, is crafted to boost data-driven methodologies for automatic humanoid rigging. 

\subsection{Automatic Rigging}

Automatic rigging consists of two tasks, skeleton construction and skinning. Skeleton construction is about setting up a hierarchical bone structure and the corresponding joint positions, according to the body anatomy and expected deformation. Skinning involves assigning  influence to skeletons over vertices in 3D models, ensuring natural deformation during skeletal animation. 

\textbf{Traditional Method.} Pinocchio \cite{baran2007automatic} employs the medial axis to position the skeletons via optimization, but cannot work well when the mesh shape deviates from the skeleton template. Geometric-based skinning methods \cite{baran2007automatic,wareham2008bone,jacobson2011bounded, kavan2012elasticity} assume skeleton influence is proportional to proximity, which can cause inaccuracies in meshes with irregular body shapes. 

\textbf{Data-driven Method.} RigNet~\cite{xu2020rignet} learns mesh offsets, attention, and bone connection probabilities, extracting joints via attention clustering and constructing skeletons with Minimum Spanning Tree. It can not support motion tasks well because the predicted skeleton has uncertain topology and lacks semantic labels. 
Based on 3D vertex features directly extracted by Graph Neural Network~(GNN) backbone, \cite{li2021learning, ma2023tarig} learn template based skeletons and \cite{liu2019neuroskinning,xu2020rignet,li2021learning,pan2021heterskinnet,mosella2022skinningnet,ma2023tarig} learn skinning weights. However, they all struggle with models that have diverse body shapes, intricate clothing and accessories, or complex face topologies. Unlike these approaches, our method effectively utilizes front view rendering to establish robust priors and employs a 2D-3D lifting technique to accurately determine skeleton positions with a coarse-to-fine approach, and overcome the skinning challenge by using a point transformer-based encoder to accurately capture vertex features in 3D space.

\section{The Basis - HumanRig Dataset}

As depicted in Fig.~\ref{fig:data pipeline}, the data acquisition pipeline for HumanRig is highly efficient and is detailed as follows.

\begin{table}[t]
  \centering
  
  \vspace{0.4em} 
  \resizebox{0.5\textwidth}{!}{ 
  \begin{tabular}{@{} l c c c c c c @{}}
    \toprule
    \textbf{Dataset} & \textbf{\#} & \textbf{Uniform Skel} & \textbf{Clothing} & \textbf{Hair} & \textbf{Bind Pose}  & \textbf{Source} \\
    \midrule
    Mixamo \cite{blackman2014rigging}       & 108   & \ding{51}   & \ding{51}   & \ding{51}  & T-pose   & artist-created \\
    SMPL \cite{loper2015smpl}               & N/A     & \ding{51}   & \ding{53}    & \ding{53}   & T-pose  & artist-created \\
    RigNetv1 \cite{xu2020rignet}        & 1729  & \ding{53}    & \ding{51}   & \ding{51}  & T-pose & artist-created \\
    HumanRig (ours)   & 11434   & \ding{51}   & \ding{51}   & \ding{51}  & T-pose   & AI-generated\\
    \bottomrule
  \end{tabular}
  }
  \caption{\textbf{Humanoid Rigging Dataset Comparisons.} For RigNetv1, only T-pose humanoid meshes picked out to compare.}
  \label{tab:dataset}
\end{table}

\subsection{3D Humanoid Models Preparation}

Constructing a large-scale 3D humanoid model dataset presents significant challenges, primarily due to the labor-intensive nature of requiring extensive input from professional artists. However, the rapid advancement of AI-based 2D image generation technology has enabled the creation of diverse humanoid character images from textual descriptions. Furthermore, leveraging AI-driven 3D model generation techniques, it has become feasible to efficiently derive high-quality 3D models from a single image.

\textbf{T-pose Image Generation.} 3D humanoid characters are commonly modeled in neutral poses like T-pose or A-pose, for enhanced mesh deformation and motion capture (mocap) data compatibility. Our work adopts the T-pose as the bind pose to streamline the rigging process for 3D humanoid models. We introduce a pose-controlled text-to-image framework for generating T-pose character images by integrating stylized text prompts with reference T-pose images. Our method includes the following steps:

\begin{enumerate}
    \item Curating 5,000 detailed text prompts for a range of characters adorned with various clothing accessories, by employing a Large Language Model~(LLM) to enhance the prompt template.
    \item Generating 50 T-pose maps with diverse head-to-body ratios by utilizing DWPose \cite{yang2023effective} to extract 2D poses from a curated set of humanoid images.
    \item Synthesizing 20,000 T-pose humanoid images by randomly selecting a text prompt and a T-pose map, and inputting them into Stable Diffusion~(SD)~\cite{rombach2022high} and the corresponding Pose ControlNet~\cite{zhang2023adding}.
    \item Filtering out low-quality images, ultimately collecting 17,268 high-quality humanoid T-pose images.
\end{enumerate}

\noindent This framework guarantees that the generated characters capture the desired style and aesthetics while maintaining accurate body proportions and proper T-pose alignment. As a result, we can produce a broad range of diverse and precisely posed T-pose humanoid character images, covering various head-to-body ratios.

\textbf{T-pose Mesh Generation.} The swift progress in 3D generation has enabled the rapid production of large-scale and high-quality 3D meshes with unprecedented speed. We utilize an Image-to-3D pipeline for generating 3D meshes, which includes open-source methods like InstantMesh \cite{xu2024instantmesh} and Unique3D \cite{wu2024unique3d}, as well as commercial tools such as Hyperhuman Rodin \cite{zhang2024clay} and Triposr-v2 \cite{tochilkin2024triposr}. By inputting meticulously prepared T-pose character images, we produce a large batch of T-pose humanoid meshes. We manually filter out low-quality ones, selecting 14,662 high-quality meshes from an initial 17,268. 

\begin{figure}
  \centering
    \includegraphics[width=1.0\linewidth]{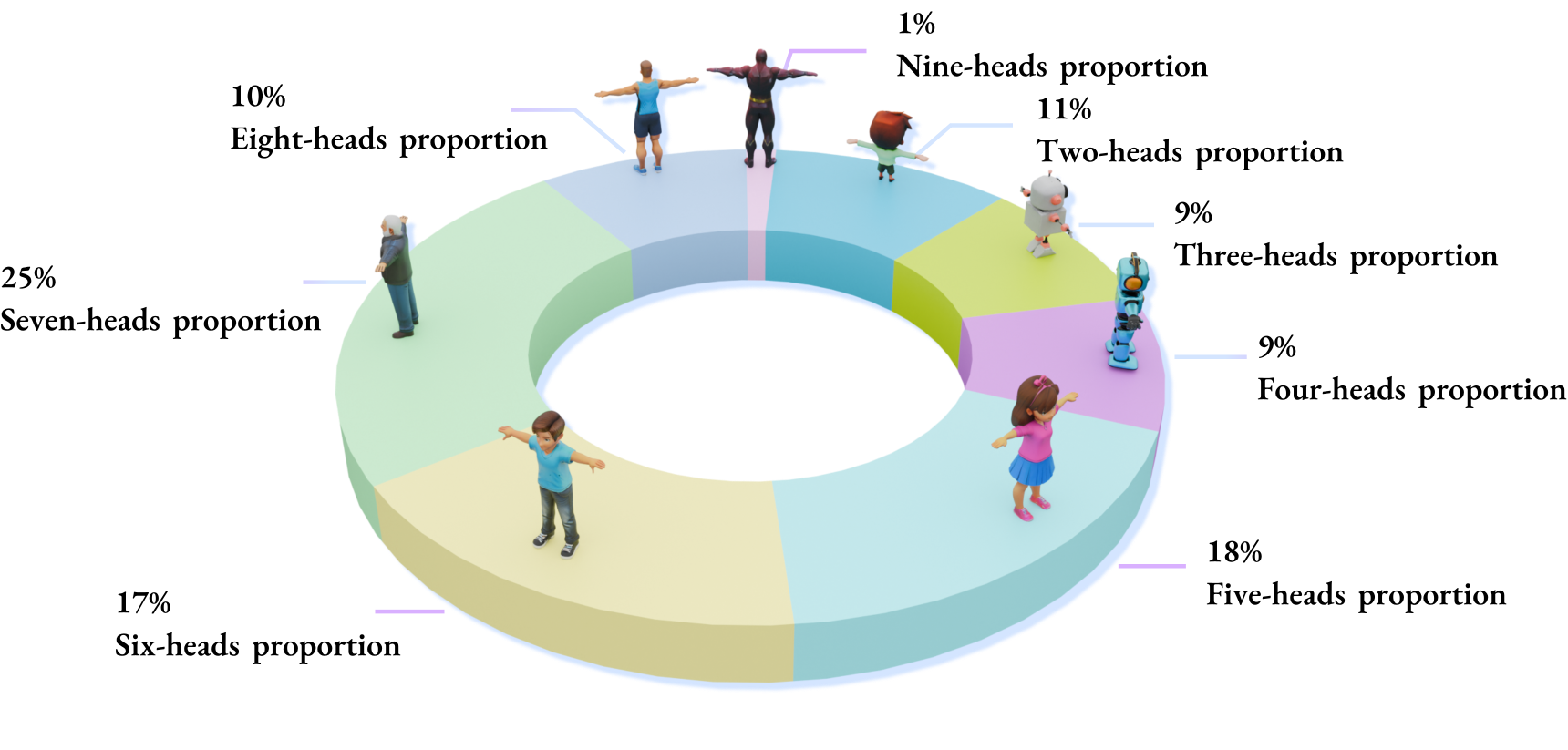}
    \caption{\textbf{Head-to-body Ratio Diversity Statistics} of HumanRig.}
    \label{fig:hbr}
\end{figure}

\begin{figure*}[ht]
  \centering
   \includegraphics[width=1.0\linewidth]{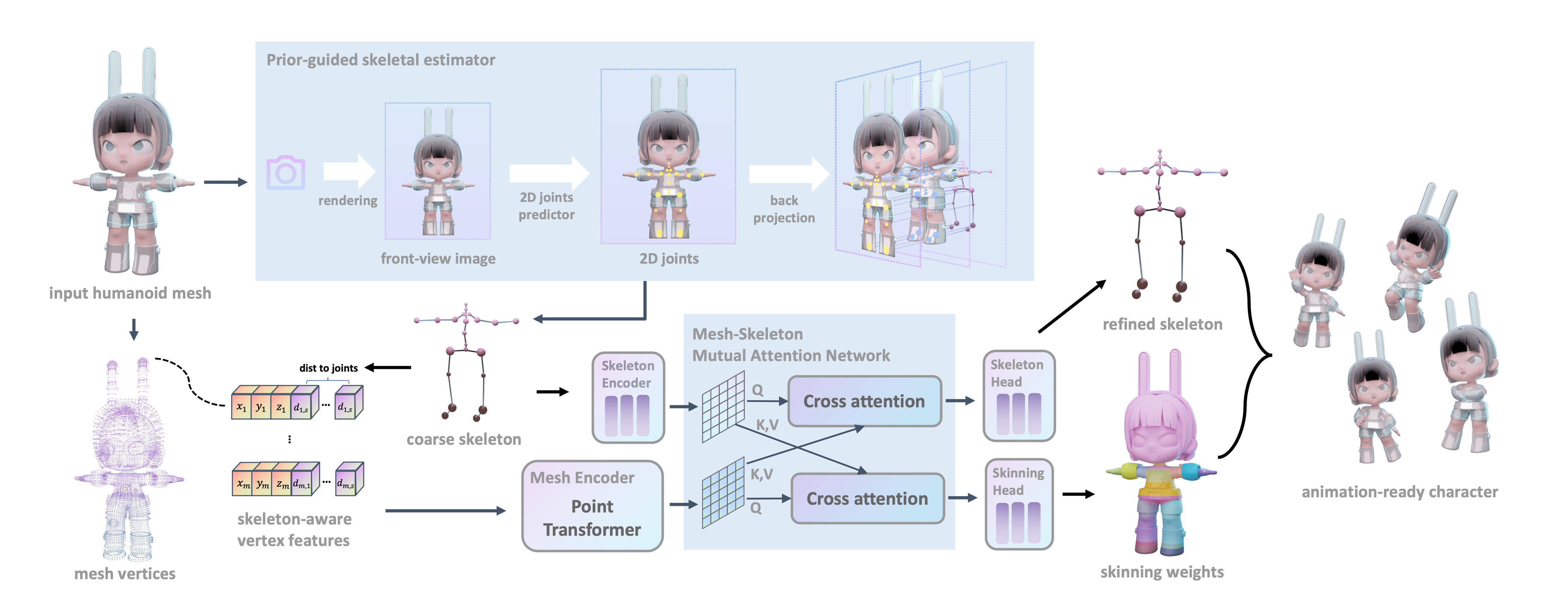}

   \caption{\textbf{Method Overview.} Given a humanoid mesh, a coarse skeleton is predicted by an Prior-guided skeleton estimator (PGSE) and helps construct the skeleton-aware vertex features. They are fed into Skeleton Encoder and Mesh Encoder, respectively, then fused by a Mesh-Skeleton Mutual Attention Network to predict a refined skeleton and skinning weights with a joint learning strategy. Finally, the skeleton and skinning weights are combined to produce the animation-ready character. }
   \label{fig:method}
\end{figure*}

\subsection{Humanoid Models Rigging }
To animate a newly created humanoid character using mocap data, it should have a practical skeleton topology that is compatible with mocap data, rather than binding to joints without semantic labels, like \cite{xu2020rignet}. To simplify the mocap-to-deformation process, we utilize the Mixamo skeleton topology, enabling our rigged character to be compatible with existing animation software and game engines. Naturally, we utilize the Mixamo semi-automated rigging tool \cite{blackman2014rigging} to speed up the annotation process. 
Poor-quality rigging data is filtered out, and artists manually refine the remaining models. Ultimately, we produce 11,434 high-quality rigged humanoid models.


\subsection{Data Postprocessing}
We normalize the prepared rigged humanoid models within a bounding box of $[-0.5, 0.5]$ to ensure proper alignment. From the rigging results, we extract 3D skeleton joint positions of size $[s, 3]$ and the skinning weight matrix of size $[m, s]$, where $s$ represents the number of skeleton joints, uniformly set to 22 in this work, and $m$ denotes the variable number of vertices across different humanoid models. Additionally, we generate a front-view image using a fixed camera. Utilizing the projection matrix derived from the camera's intrinsic and extrinsic parameters, we project the 3D skeleton joints onto the front-view image, thereby obtaining the corresponding 2D skeleton joint positions. 

\subsection{Data Summary }
As presented in Table \ref{tab:dataset}, we conduct a comparative analysis of our dataset with existing rigged humanoid datasets. Existing datasets are limited in quantity, diversity or skeleton unity. Our dataset bridges these gaps by offering a large-scale collection of 11,434 AI-generated meshes with diverse head-to-body ratios, aligned with a consistent Mixamo skeleton, thereby enhancing both the quantity and diversity of available models for rigging research and applications. Fig.~\ref{fig:hbr} illustrates models within our dataset, showcasing the diverse head-to-body ratios and their corresponding data proportions. Each annotated sample comprises a rigged T-pose humanoid mesh, 3D skeleton joint positions, a skinning weight matrix, a front-view image, and the associated camera parameters along with the 2D skeleton joint positions.

\section {Automatic Rigging}
Capitalizing on the expansive HumanRig dataset characterized by a uniform skeleton topology, we propose a data-driven automatic rigging framework. As depicted in Fig.~\ref{fig:method}, our approach begins by initializing a coarse skeleton using a Prior-Guided Skeleton Estimator~(PGSE). The skeleton serves a dual purpose: It facilitates the learning of joint positions from coarse to fine, and it enhances vertex attributes to yield skeleton-aware vertex features, thereby simplifying the learning process for both the skeleton and skinning weights.
Furthermore, we employ two encoders to extract skeleton and mesh features respectively: a MLP-based Skeleton Encoder for extracting skeleton features and a Point Transformer-based Mesh Encoder for extracting mesh features. These distinct features are then cross-integrated through a Mesh-Skeleton Mutual Attention Network~(MSMAN) to achieve deeper feature integration. We will introduce each of these modules individually.

\subsection{Prior-guided Skeleton Estimator}
The Prior-Guided Skeleton Estimator~(PGSE) utilizes 2D skeleton priors from the model's front view and projects them into approximate 3D positions using camera parameters. This approach significantly simplifies the task of learning skeleton positions directly from 3D mesh geometry by providing an initial estimation, which can then be refined using mesh data. Specifically, we fine-tune RTMPose~\cite{jiang2023rtmpose} on the front view image to predict 2D template joints $P_{j2D} \in \mathbb{R}^{s \times 2}$ accurately. Then, each 2D joint $\mathbf{J}_{2D}$ is back-projected into 3D as a ray:
$$
\tilde{\mathbf{X}_r}\left(\mu ; \mathbf{J}_{2D}\right)=\mathrm{P}_c \tilde{\mathbf{J}}_{2D}+\mu \tilde{\mathbf{X}}_c
$$
where $\mathrm{P}_c \in \mathbb{R}^{4 \times 3}$ is the pseudo-inverse of the camera projection matrix, $\mathbf{X}_c$ is the 3D location of the camera center, and the symbol with superscript tilde denotes the corresponding homogeneous coordinate.
Finally, we determine the coarse 3D skeleton $P_{skec} \in \mathbb{R}^{s \times 3}$ by calculating the intersections of each ray with the mesh surface and using the midpoints of the first and last intersections for each 3D coarse joint.

\subsection{Skeleton and Mesh Encoder}
We utilize a Skeleton Encoder composed of a simple 3-layer Multi-Layer Perceptron (MLP) to process the coarse 3D skeleton points and extract skeleton features $f_{s} \in \mathbb{R}^{s \times c}$, where $c$ denotes the number of channels. This encoder is efficient due to the skeleton's fixed topology and limited number of joints.

For mesh feature extraction, we begin by extracting all vertices and constructing skeleton-aware vertex features $f_{v} \in \mathbb{R}^{m \times (3+s)}$, where each of the $m$ vertices has 3 positional coordinates and $s$ euclidean distance features relative to $P_{skec}$. We then employ a U-shaped Point Transformer~\cite{zhao2021point} as the Mesh Encoder to extract deep mesh features, with the output of the last up-transition layer serving as our encoder output $f_{m} \in \mathbb{R}^{m \times c}$. The Point Transformer-based Mesh Encoder effectively integrates local geometric correlation features with global geometric semantic features. Compared to commonly used GNN-based Mesh Encoders, it exhibits superior generalization for AI-generated meshes that often have irregular face topology and incorporate a large number of vertices.

\subsection{Mesh-Skeleton Mutual Attention Network}
Mesh-skeleton Mutual Attention Network (MSMAN) is further proposed to integrate Skeleton features $f_{s}$ and Mesh features $f_{m}$ through mutual cross-attention mechanisms. The skeleton features provide mesh features with body part affiliation, which aids in the semantic understanding of mesh vertices and enhances the accuracy of skinning weight predictions. In a similar vein, mesh features bolster the local feature representation of the skeleton, which is beneficial for refining the precise positions of the skeleton. Taking one side as an example, we introduce a multi-head cross-attention mechanism to enhance mesh features, which is formalized as follows: 
\begin{equation}
Q = {f_s}; \quad K = {f_m}; \quad V = {f_m}
\end{equation}

\begin{equation}
f_{m \to s} =  \text{softmax} \left( \frac{QK^T}{\sqrt{d}} \right) V       
\end{equation}
where $f_{m \to s} \in \mathbb{R}^{s \times {c}}$ is the enhanced skeleton features.

For the other side of the interaction, we adopt the same formulation but set $Q ={f_m}; \quad K = {f_s}; \quad V = {f_s}$ to integrate the skeleton features into the mesh features. Finally, we get $f_{s \to m} \in \mathbb{R}^{m \times {c}}$, which represents skeleton-aware mesh features. 

\begin{table*}[htbp]
  \centering
  
   \resizebox{\textwidth}{!}{ 
     
    \begin{tabular}{c|ccccc|ccccc}
    \toprule
    \multirow{2}[1]{*}{Train Set} & \multicolumn{5}{c|}{RigNetv1-human(Test)} & \multicolumn{5}{c}{HumanRig(Test)} \\
          & CD-J2J↓ & CD-J2B↓ & CD-B2B↓ & Prec↑ & L1↓   & CD-J2J↓ & CD-J2B↓ & CD-B2B↓ & Prec↑ & L1↓ \\
    \midrule
    RigNetv1-human & 0.0274 & 0.0258 & 0.0208 & 0.7445 & 0.2573 & 0.0163 & 0.0155 & 0.0126 & 0.8364 & 0.1700 \\
    HumanRig-small & 0.0189 & 0.0173 & 0.0110 & 0.6502 & 0.3325 & 0.0054 & 0.0050 & 0.0028 & 0.8371 & 0.0789 \\
    HumanRig & \textbf{0.0071} & \textbf{0.0064} & \textbf{0.0033} & \textbf{0.7452} & \textbf{0.1564} & \textbf{0.0027} & \textbf{0.0025} & \textbf{0.0012} & \textbf{0.9271} & \textbf{0.0610} \\
    \bottomrule
    \end{tabular}%

    }
    \caption{\textbf{Cross-dataset comparisons} of RigNetv1-human and HumanRig.}
  \label{tab:dataseteffect}%
\end{table*}%

\subsection{Rigging Head and Training Loss}
We employ two separate MLP-based heads to independently predict skeleton positions $P_{ske} \in \mathbb{R}^{s \times {3}}$ and skinning weights $P_{skin} \in \mathbb{R}^{m \times {s}}$. For skeleton joints prediction, we use Mean Squared Error (MSE) Loss:
\begin{equation}
\mathcal{L}_{skeleton}= \frac{1}{s} \sum_{i=1}^s \left(P_{{ske}_{i}}-G_{{ske}_{i}}\right)^2
\end{equation}
Skinning prediction can be conceptualized as optimizing the selection probabilities of various skeleton joints for each mesh vertex. Consequently, we employ the Kullback-Leibler divergence loss to minimize the distance between the predicted skin weight distribution $P_{skin}$ and the ground truth distribution $G_{skin}$:
\begin{equation}
\mathcal{L}_{skinning} =\frac{1}{m} \sum_{i=1}^m \sum_{j=1}^s P_{skin_{i j}}\left(\log \frac{P_{skin_{i j}}}{G_{skin_{i,j}}}\right)
\end{equation}
The total loss is constructed as:
\begin{equation}
    \mathcal{L}_{total} = \mathcal{L}_{skinning} + \mathcal{L}_{skeleton}
\end{equation}

\section{Experiments}

\begin{table}[htbp]
  \centering
  \resizebox{0.5\textwidth}{!}{ 
    \begin{tabular}{cccccc}
    \toprule
    Method & CD-J2J↓ & CD-J2B↓ & CD-B2B↓ & Prec↑ & L1↓ \\
    \midrule
    Ours(w/o PGSE) & 0.0110 & 0.0061 & 0.0031 & 0.8363 & 0.1630 \\
    Ours(w/o MSMAN) & 0.0080 & 0.0076 & 0.0039 & 0.9214 & 0.0630 \\
    Ours(complete) & \textbf{0.0027} & \textbf{0.0025} & \textbf{0.0012} & \textbf{0.9271} & \textbf{0.0610} \\
    \bottomrule
    \end{tabular}%
    }
     \caption{\textbf{Architecture Study} that shows the influence of PGSE and MSMAN.}
  \label{tab:ablation}%
\end{table}%

\begin{table}[htbp]
  \centering
  \resizebox{0.5\textwidth}{!}{ 
    \begin{tabular}{cccccc}
    \toprule
    Mesh Encoder & CD-J2J↓ & CD-J2B↓ & CD-B2B↓ & Prec↑ & L1↓ \\
    \midrule
    GraphSAGE~\cite{hamilton2017inductive} & 0.0031 & 0.0030 & 0.0016 & 0.8153 & 0.0764 \\
    GraphTransformer~\cite{shi2020masked} & 0.0030 & 0.0029 & 0.0016 & 0.8100 & 0.0760 \\
    GMEdgeNet~\cite{xu2020rignet} & 0.0028 & 0.0026 & 0.0014 & 0.8290 & 0.0751 \\
    Ours(Point Transformer) & \textbf{0.0027} & \textbf{0.0025} & \textbf{0.0012} & \textbf{0.9271} & \textbf{0.0610} \\
    \bottomrule
    \end{tabular}%
    }
    \caption{\textbf{Mesh Encoder Study} that shows our superior performance compared to GNNs.}
  \label{tab:meshencoder}%
\end{table}%

\textbf{Implementation details.} Our framework is trained in two stages. First, we use the mmpose~\cite{mmpose2020} framework to train a 2D joint predictor based on RTMPose~\cite{jiang2023rtmpose}. We then apply PGSE to extract coarse skeletons from input meshes, which are used in the second stage to train other components. Both stages are trained on our HumanRig dataset, with the same 80\%-10\%-10\% data split. The training is performed on 2 NVIDIA RTX A6000 GPUs using the AdamW~\cite{loshchilov2017decoupled} optimizer and a MultiStepScheduler. The training lasts for 300 epochs, with the learning rate halved after 50 epochs, starting from an initial rate of 1e-3 and a batch size of 16.

\textbf{Evaluation metrics.} For skeleton construction evaluation, we employ the metrics CD-J2J, CD-J2B and CD-B2B proposed by \cite{xu2020rignet}. For skinning evaluation, we adopt the same metrics as \cite{liu2019neuroskinning,xu2020rignet,pan2021heterskinnet,mosella2022skinningnet}, including skinning precision and L1-norm. Interested readers are referred to the original papers for detailed definitions of these metrics. For deformation quality, we evaluate deformation error by calculating the average Euclidean distance between deformed vertex positions and ground truth ones, using 10 random poses with joint rotations within a range of ±10 degrees.

\subsection{HumanRig Dataset Study}

To assess the effectiveness of our HumanRig dataset, we perform a comparative study with RigNetv1.
RigNetv1, the largest rigging dataset, lacks a uniform skeleton and consistent initial poses. To ensure a fair comparison, we select 1,729 T-pose humanoid meshes, rigged them with the Mixamo skeleton via our annotation pipeline, and labeled them as RigNetv1-human.
We train our HumanRig method on the training sets of RigNetv1-human and HumanRig respectively, and subsequently evaluated both models on both test sets. Additionally, to neutralize the effect of dataset size, we created HumanRig$_{small}$, a random subset of HumanRig comprising 1,729 samples.
Table \ref{tab:dataseteffect} demonstrate valuable insights. First, a model trained on HumanRig$_{small}$ surpasses one trained on RigNetv1-human on the HumanRig test set, even outperforming the intra-dataset training of RigNetv1-human in skeleton construction task. Second, expanding the dataset size to the full HumanRig dataset significantly improves performance in both tasks. Lastly, the model trained on our dataset excels on both artist-created and AI-generated models.

\begin{figure}
  \centering
    \includegraphics[width=1.0\linewidth]{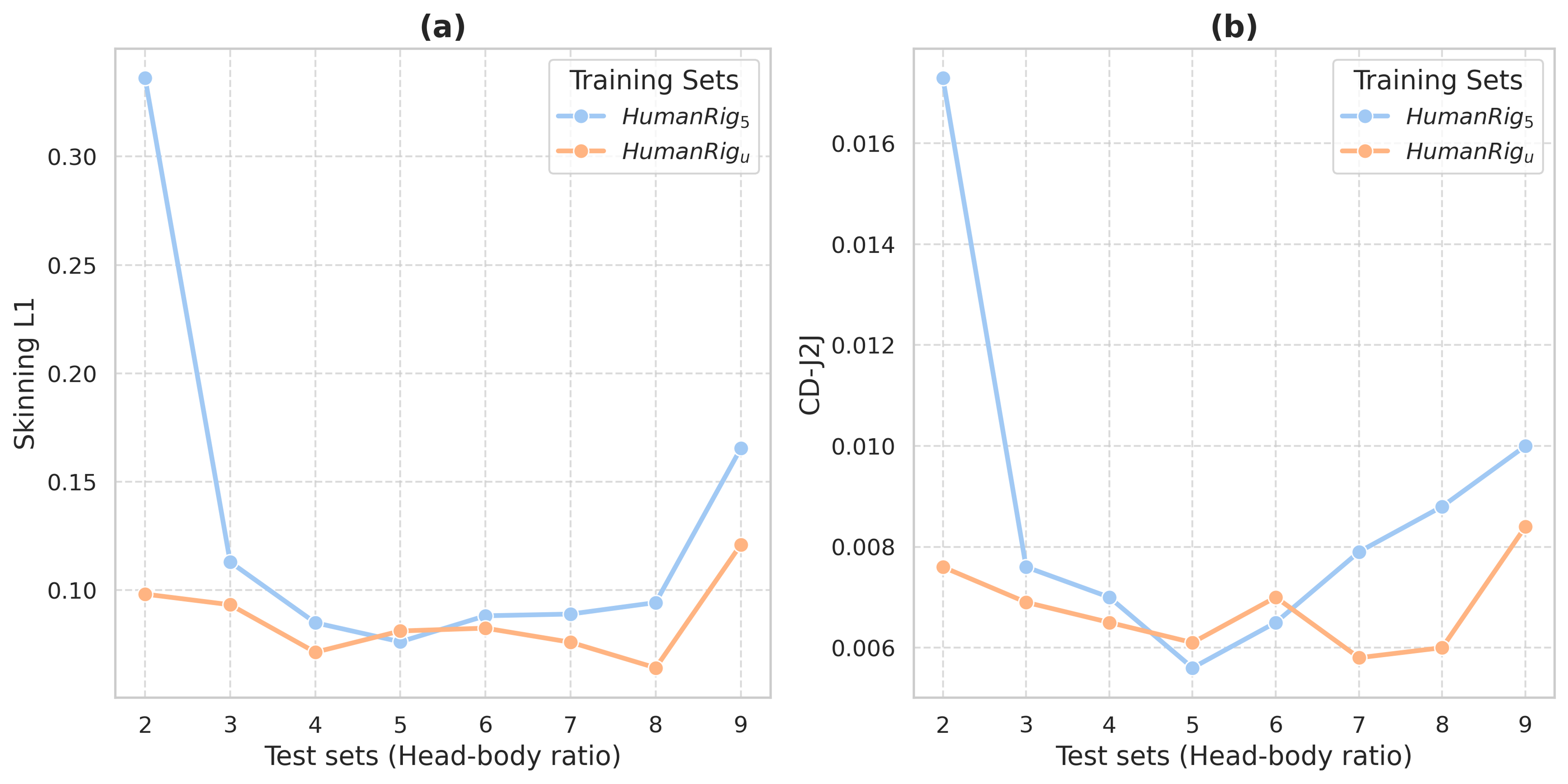}
    \caption{\textbf{Study on the Importance of Diversity in Body Shapes.} }
    \label{fig:hbre}
\end{figure}

In addition to the large scale, the superior performance is also attributed to the diverse body proportions (see Fig. \ref{fig:hbr}). 
We underscore its importance by an experiment that categorizes the HumanRig training and test sets into subsets based on head-body ratios, with the subset containing five heads designated as HumanRig$_5$. We then create a new, balanced subset, HumanRig$_u$, by evenly sampling from all subsets to match the size of HumanRig$_5$. 
We train two HumanRig models separately on HumanRig$_5$ and HumanRig$_u$, and then evaluate them on test subsets ranging from two-heads to nine-heads. 
The experimental results in Fig.~\ref{fig:hbre} indicate that for the model trained on HumanRig$_5$, the performance declines as the head-body ratio deviates from the five heads in both tasks. When considering various head-to-body ratios, it exhibits a more balanced performance distribution and overall improvement.

\begin{figure*}[ht]
  \centering
   \includegraphics[width=1.0\linewidth]{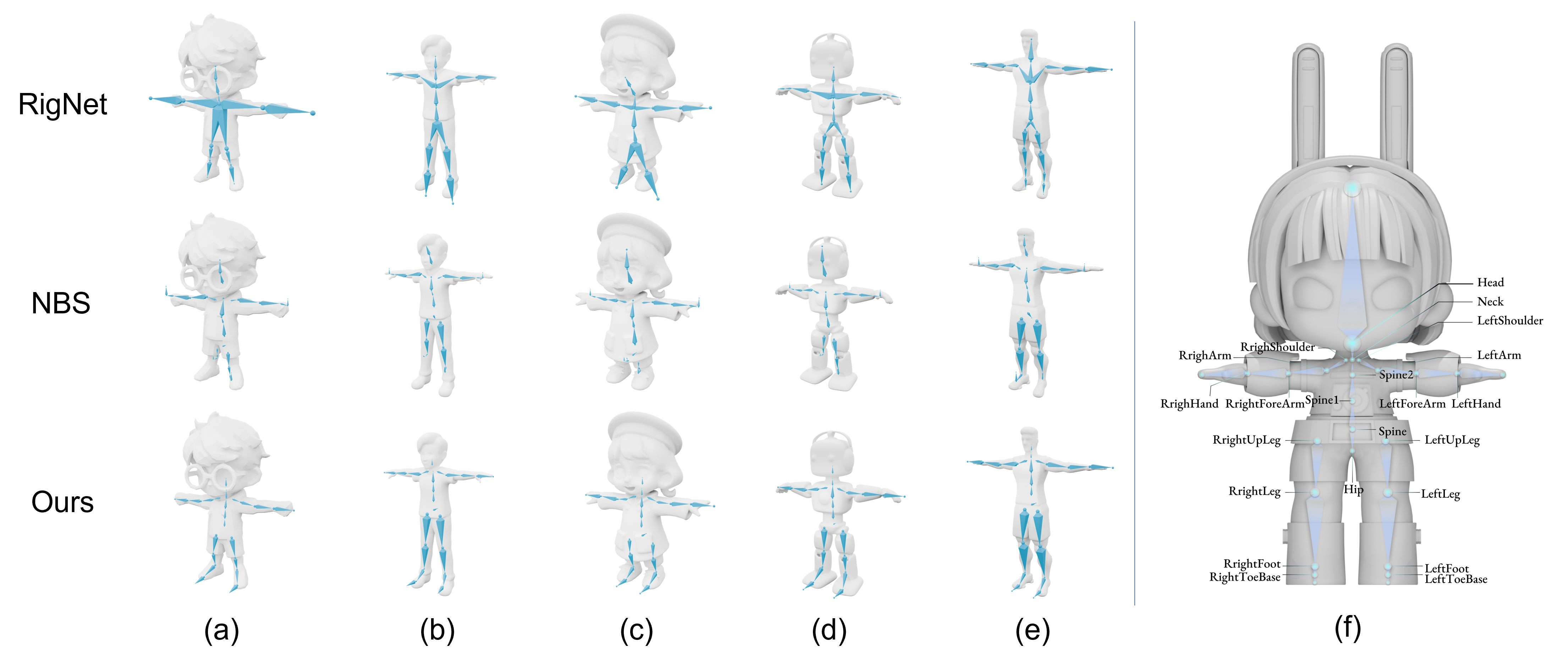}
   \caption{\textbf{Skeleton construction comparisons.} RigNet outputs irregular skeletons, which presents challenges to motion applications. NBS performs poorly in joint position prediction. Our method achieves the most plausible humanoid skeletons according to different human shapes from character (a) to (e). The skeleton topology we utilized is illustrated in character (f).}
   \label{fig:rigsota}
\end{figure*}

\subsection{Architecture Ablation Study}

\textbf{PGSE Design.} After eliminating PGSE, the network directly regresses joint positions from mesh features without the coarse skeleton's guidance. This also limits the mesh encoder's input to vertex coordinates. Table \ref{tab:ablation} demonstrates that PGSE significantly reduces the difficulty of skeleton construction (CD-J2J drops from 0.0110 to 0.0027). Additionally, skeleton-aware vertex features substantially improves skinning performance.

\textbf{MSMAN Design.} The MSMAN is integral to our method as it enables the joint optimization of skeleton construction and skinning. To isolate the effects of MSMAN, we train an expert network that estimates skinning weights directly from the mesh encoder's output, and use the coarse skeleton as the final skeleton output. Table \ref{tab:ablation} illustrates that the coarse skeleton itself achieves commendable accuracy, but its refinement by MSMAN still results in a notable improvement. Furthermore, skinning performance is improved by the joint optimization. These findings underscore the significance of MSMAN's mutual attention mechanism in effectively integrating mesh features and skeletal representations within a high-level semantic space.

\textbf{Mesh Encoder Design.} In this work, we deviate from the traditional approach of utilizing GNNs for mesh feature extraction by introducing a U-shaped Point Transformer. To investigate the impact of this design choice, we replace our mesh encoder with GMEdgeNet proposed by RigNet, and substitute its EdgeConv \cite{wang2019dynamic} operator with other widely-used GNNs such including GraphSAGE \cite{hamilton2017inductive} and GraphTransformer \cite{shi2020masked}. The results presented in Table \ref{tab:meshencoder} demonstrate that our mesh encoder achieves superior performance. This improvement is due to the more comprehensive and efficient aggregation of mesh vertex features based on 3D spatial distances, compared to aggregation along surface edges in GNNs. Additionally, edges in AI-generated 3D models are often irregular and lack semantic information, thus, a mesh encoder that ignores edge information exhibit improved generalization and robustness.

\subsection{Comparisons}

Due to the lack of effective datasets, there are few data-driven methods~\cite{liu2019neuroskinning,xu2020rignet,li2021learning,pan2021heterskinnet,mosella2022skinningnet,ma2023tarig} for studying humanoid neural rigging. Unfortunately, as far as we know, only RigNet \cite{xu2020rignet} and NBS \cite{li2021learning} have been open-sourced. We conduct the following experiments to demonstrate the superiority of our method.

\begin{figure*}[ht]
  \centering
   \includegraphics[width=0.87\linewidth]{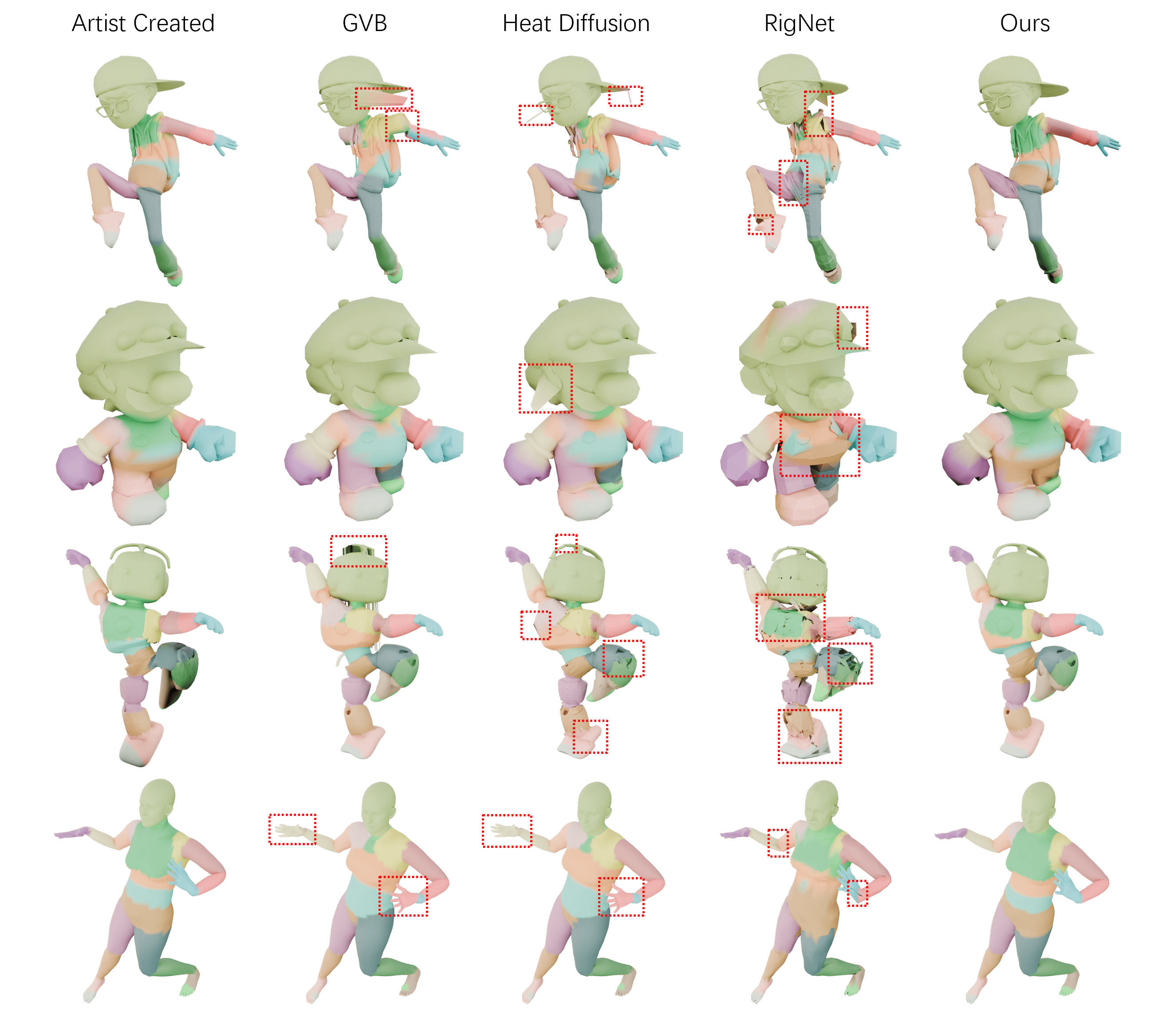}
   \caption{\textbf{Skinning and Deformation Comparisons.} Different colors are assigned to each vertex according to the most influential joint, determined by skinning weights. Previous methods struggle with meshes that have special body shapes or intricate accessories (red box), while our method achieves superior deformation quality. \textbf{More intuitive results are visualized in the supplementary video.}}
   \label{fig:skinsota}
\end{figure*}

\textbf{Skeleton Construction.} For skeleton prediction, we compare our method with RigNet \cite{xu2020rignet} and NBS \cite{li2021learning}, as illustrated in Fig.~\ref{fig:rigsota}.
RigNet results in a lack of different key joints in all samples because it does not incorporate a humanoid skeleton prior, leading to uncertain joint placements and bone connections without semantic context. Although NBS utilizes a fixed skeleton template of SMPL \cite{loper2015smpl}, it produces inaccurate joint positions, especially in the lower limbs. This inaccuracy stems from the challenge of directly regressing joint positions from mesh features. Moreover, as it exclusively trained on the SMPL dataset, it struggles to generalize well to diverse meshes that exhibit varying head-to-body ratios. In contrast, by leveraging a robust 2D prior, we predict joints based on a standard skeleton template, resulting in more plausible skeletons.

\begin{table}[t]
  \centering
  \resizebox{0.5\textwidth}{!}{ 
   \begin{tabular}{ccccc}
    \toprule
    Test set & Heat Diffusion~\cite{baran2007automatic} & GVB~\cite{dionne2014geodesic}   & RigNet~\cite{xu2020rignet} & Ours \\
    \midrule
    RigNetv1-human & 0.00405 & 0.00306 & 0.00252 & \textbf{0.00198} \\
    HumanRig & 0.00236 & 0.00224 & N/A   & \textbf{0.00084} \\
    \bottomrule
    \end{tabular}%
    }
    \caption{\textbf{Deformation Error Study.} RigNet is not evaluated on HumanRig as it only handles 1k-5k vertex meshes. Our method achieves the best deformation quality on both test sets.}
  \label{tab:deformation}%
\end{table}%

\textbf{Skinning and Deformation.}  We conducted a comparative analysis with Heat Diffusion~\cite{baran2007automatic}, Geodesic Voxel Binding (GVB)~\cite{dionne2014geodesic}, and RigNet~\cite{xu2020rignet}. These methods can predict skinning weights given an input mesh and its corresponding skeleton. Heat Diffusion and GVB are traditional geometry-based methods, whereas RigNet employs a data-driven approach by training an independent skinning network. For fair comparison, we input them the same skeletons predicted by our HumanRig method. Table \ref{tab:deformation} presents the quantitative assessment of deformation errors. RigNet can not handle complex meshes with over 5K vertices, while our approach outperforms on both artist-created and AI-generated meshes. Additionally, Fig.~\ref{fig:skinsota} provides a qualitative visualization, which indicates that our method excels in maintaining more lifelike and fluid animations across varying head-to-body ratios, intricate clothes or accessories. This benefits from the strong point-transformer-based mesh encoder, which competently differentiates between distinct body parts to generate more precise skinning weights.

\section{Conclusions}

In this work, we have addressed the critical need for a comprehensive dataset and a robust framework for automatic rigging of 3D humanoid character models. We introduce HumanRig, a large-scale dataset of AI-generated T-pose humanoid models, all rigged with a consistent skeleton topology, significantly surpassing the previous datasets in terms of size, diversity, complexity and practical motion applications.
Leveraging the HumanRig dataset, we propose a novel automatic rigging framework that integrates a Prior-Guided Skeleton Estimator (PGSE), a U-shaped Point Transformer-based Mesh Encoder and a Mesh-Skeleton Mutual Attention Network (MSMAN). Our approach excels in skeleton construction and skinning prediction, effectively managing challenges from irregular shapes, detailed clothing or accessories, and differing head-to-body ratios. Compared to GNN-based approaches, our method provides superior generalization to complex, AI-generated meshes.

Our method still have limitations. First, the skeleton template has not considered small body parts such as fingers, which requires generating finer meshes in datasets. Second, it is exciting if our method could be extended to the quadruped or any objects.



\newpage
{
    \small
    \bibliographystyle{humanrig}
    \bibliography{humanrig}
}


\end{document}